\documentclass[10pt,twocolumn,letterpaper]{article}
\usepackage[pagenumbers]{wacv} 
\usepackage{graphicx}
\usepackage{amsmath}
\usepackage{amssymb}
\usepackage{booktabs}

\usepackage{paralist} 
\usepackage[font=small,labelfont=bf]{caption}
\usepackage{subcaption}
\usepackage{verbatim}
\usepackage{float}
\DeclareUnicodeCharacter{2208}{\in} 
\usepackage{algorithm}
\usepackage{algorithmic}
\usepackage{booktabs}

\usepackage{comment}
\usepackage[colorlinks=true, linkcolor=red, urlcolor=red]{hyperref}
\usepackage[capitalize]{cleveref}
\crefname{section}{Sec.}{Secs.}
\Crefname{section}{Section}{Sections}
\Crefname{table}{Table}{Tables}
\crefname{table}{Tab.}{Tabs.}

\def\wacvPaperID{1914} 
\def\confName{WACV}
\def\confYear{2025}

\begin{document}

\title{Reversing the Damage: A QP-Aware Transformer-Diffusion Approach for 8K Video Restoration under Codec Compression}

\author{
\begin{tabular}{ccc}
    Ali Mollaahmadi Dehaghi & Reza Razavi & Mohammad Moshirpour \\
    {\small University of Calgary} & {\small Userful Corporation} & {\small University of California, Irvine} \\
    {\small Calgary, AB, Canada} & {\small Calgary, AB, Canada} & {\small Irvine, CA, USA} \\
    {\tt\small ali.mollaahmadidehag@ucalgary.ca} & {\tt\small reza.razavi@userful.com} & {\tt\small mmoshirp@uci.edu}
\end{tabular}
}

\maketitle

\begin{abstract}
   In this paper, we introduce DiQP; a novel Transformer-Diffusion model for restoring 8K video quality degraded by codec compression. To the best of our knowledge, our model is the first to consider restoring the artifacts introduced by various codecs (AV1, HEVC) by Denoising Diffusion without considering additional noise. This approach allows us to model the complex, non-Gaussian nature of compression artifacts, effectively learning to reverse the degradation. Our architecture combines the power of Transformers to capture long-range dependencies with an enhanced windowed mechanism that preserves spatio-temporal context within groups of pixels across frames. To further enhance restoration, the model incorporates auxiliary "Look Ahead" and "Look Around" modules, providing both future and surrounding frame information to aid in reconstructing fine details and enhancing overall visual quality. Extensive experiments on different datasets demonstrate that our model outperforms state-of-the-art methods, particularly for high-resolution videos such as 4K and 8K, showcasing its effectiveness in restoring perceptually pleasing videos from highly compressed sources. \footnote{\href{https://github.com/alimd94/DiQP}{https://github.com/alimd94/DiQP}}
\end{abstract}

\section{Introduction}
\label{sec:intro}

\begin{figure*}[ht]
  \centering
  \includegraphics[width=\textwidth]{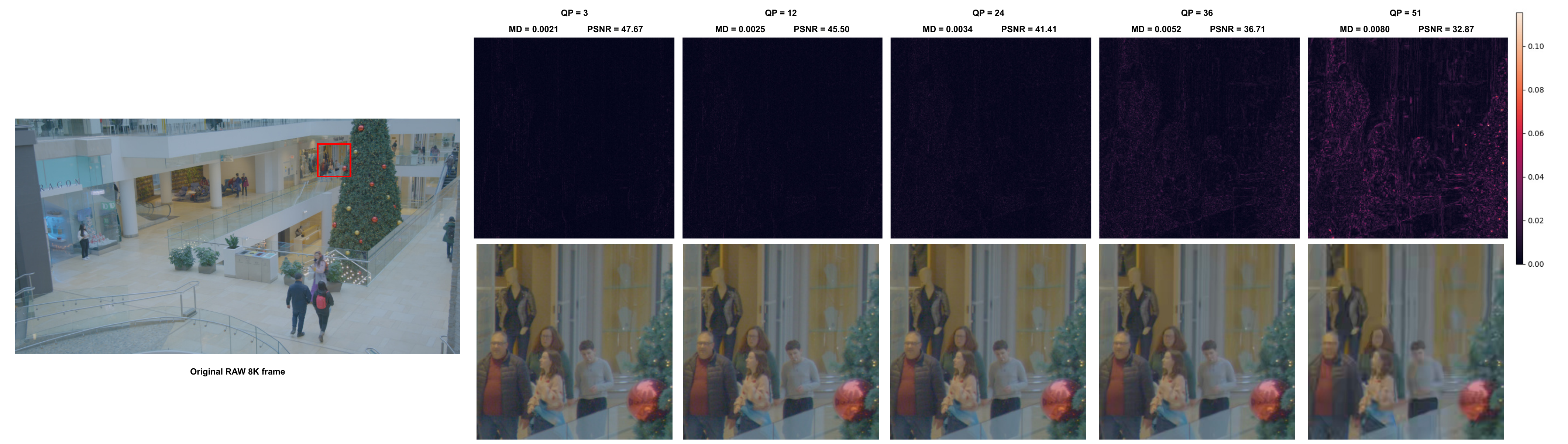}
  \caption{As the QP increases, the overall quality decreases, leading to more noticeable artifacts. Pixels are affected differently relative to each other. We calculated the Mean Absolute Difference and PSNR for these four quality levels and also present a heatmap of the affected areas. }
  \label{fig:diff}
\end{figure*}

8K video offers exceptional resolution, contrast, and motion quality, but it demands significant data and computational power for transmitting and coding \cite{Yuan2023}. With an estimated 15\% of global electricity consumption attributed to information and communication technology (ICT) by 2040 \cite{Marks2022}, and video traffic accounting for 82\% of global Internet traffic in 2022, efficient storage and transmission are increasingly crucial, particularly in bandwidth-limited scenarios prevalent in certain regions and demographics \cite{s21134589}. Video codecs offer a solution by compressing video data, but this often introduces visual artifacts like blockiness, blurring, or ringing, due to the lossy nature of compression algorithms \cite{Deng2020}. A key parameter controlling this trade-off is the Quantization Parameter (QP), which determines the level of quantization applied to transform coefficients in the video data \cite{Wu2014} (Fig. \ref{fig:diff}).

With the help of Video Restoration we can address this issue by recovering high-quality video sequences from their degraded compressed counterparts \cite{Rota2023}. Video restoration, particularly for heavily compressed videos, is a highly challenging and ill-posed problem due to the inherent trade-off between compression and quality \cite{NEURIPS2022_c573258c}. This process requires a range of techniques, including denoising \cite{9710574} to remove unwanted artifacts, deblurring \cite{Su2017} to sharpen the frame, super-resolution \cite{Chan2022,Liang2024,Liang2022} to enhance details, and crucially reducing compression artifacts – all aimed at recovering lost visual information. This task becomes even more complex when considering high-resolution videos like 8K, where the sheer volume of data exacerbates the challenges of artifact removal and quality restoration.

\begin{figure*}[ht]
  \centering
  \includegraphics[width=0.9\textwidth]{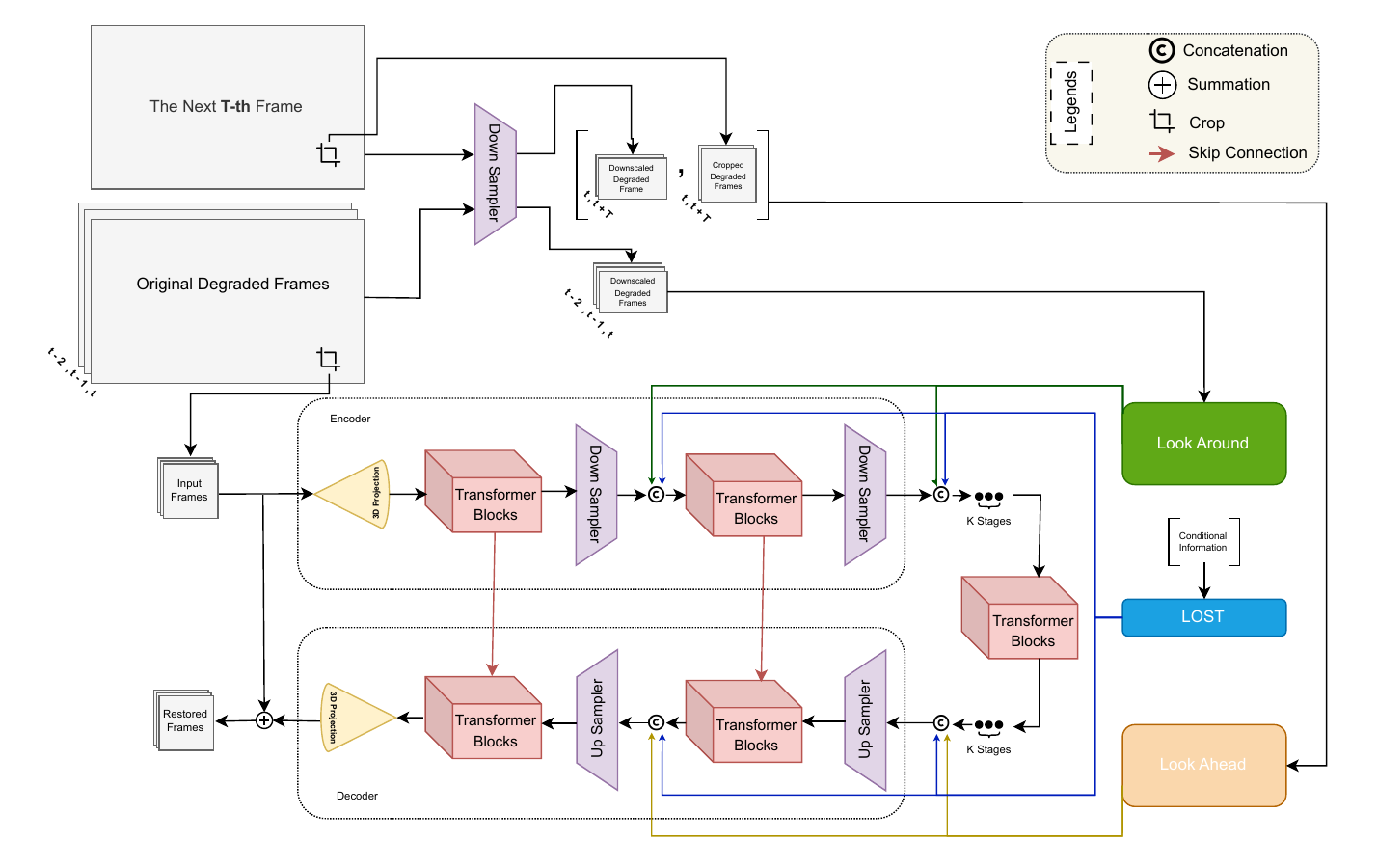}
  \caption{The overall architecture of the proposed model}
  \label{fig:main}
\end{figure*}

Our proposed model, \emph{DiQP} (Fig. \ref{fig:main}), not only introduces a novel approach to reducing video compression artifacts, but also, according to our research, is the first model specifically designed and trained for 8K videos. \emph{DiQP} uniquely reverses the codec side effects by using Denoising Diffusion \cite{Ho2020,song2021denoising}. While modern codecs such as AV1 and HEVC utilize adaptive QPs, we focus on fixed QPs to ensure robustness across varying compression levels. Unlike previous methods that add artificial noise \cite{Zhou2022}, \emph{DiQP} directly addresses complex compression artifacts by leveraging the inherent noise introduced during compression. \emph{DiQP} features a U-shaped hierarchical network inspired by \cite{Wang2022}, with skip connections and enhanced windowed self-attention for capturing long-range dependencies in high-resolution videos while preserving local context. Look Ahead and Look Around modules further enhance temporal coherence and global awareness, while LOST embedding effectively incorporates conditional data. This combination of components allows \emph{DiQP} to effectively reverse compression degradation, and significantly improve video quality restoration; particularly for 8K content.

\section{Related Works}

{\bf Video Diffusion Models'} applications contain a wide scope of video analysis tasks \cite{Xing2023}, including video generation \cite{Ho2022,Luo2023,Singer2022}, and video editing \cite{ZichengZhang2023,Qi2023}. The methodologies for these tasks share similarities, often formulating the problems as Diffusion generation tasks or utilizing the potent controlled generation capabilities of Diffusion models for downstream tasks \cite{Xing2023}. In video enhancement and restoration, CaDM \cite{Zhou2022} introduces a new approach to video streaming, reducing bitrates while improving video restoration quality compared to existing methods. This is achieved by reducing frame resolution and color depth during encoding, and then utilizing a Diffusion-based restoration process at the decoder that is aware of these encoding conditions. LDMVFI \cite{Danier2023} marks a advancement in video frame interpolation by utilizing a conditional latent Diffusion model. It features an autoencoding network tailored for video frame interpolation, incorporating efficient self-attention mechanisms and deformable kernel-based synthesis for superior performance. VIDM \cite{Chang2023} leverages a pre-trained Latent Diffusion Model (LDM) \cite{Rombach2022} to tackle video in-painting, demonstrating the adaptability of this tool. By providing a mask for first-person perspective videos, VIDM harnesses the image completion capabilities of LDM to generate seamless in-painted videos.

{\bf Video Transformers} have found applications in various domains due to their ability to model long-range dependencies efficiently  \cite{Dosovitskiy2021,Liu2021,Girdhar2019}. These applications showcase the versatility and effectiveness of Video Transformers in various video processing tasks \cite{Selva_2023}. In the restoration task, VRT \cite{Liang2024}, a new Video Restoration Transformer, allows for parallel processing of long video sequences and models long-range dependencies for video restoration. VRT jointly extracts, aligns, and fuses features at multiple scales using a novel mutual attention mechanism, achieving great performance in various video restoration tasks. RVRT \cite{Liang2022}, a recurrent video restoration transformer, combines the strengths of parallel and recurrent methods for efficient and effective video restoration. It processes video clips jointly, utilizes a larger hidden state to alleviate information loss, and introduces a novel guided deformable attention mechanism for accurate video clip alignment.

{\bf Video Restoration} has gained significant attention in recent years. FTVSR \cite{Qiu2022} uses frequency-based patch representations and attention mechanisms to address the challenges of compressed video restoration. This approach preserves high-frequency details and leverages low-frequency information to guide high-frequency texture generation, effectively reducing compression artifacts. BasicVSR++ \cite{Chan2022} improves video super-resolution using two main techniques: second-order grid propagation, which allows for more flexible information flow and aggregation across frames, and flow-guided deformable alignment, which utilizes optical flow to refine feature alignment across misaligned frames. These enhancements lead to better utilization of spatio-temporal information and improve overall performance. CAVSR \cite{YingweiWang2023} is designed to enhance video super-resolution specifically for compressed videos. It incorporates a compression encoder to assess compression levels in frames, using metadata such as frame type and motion vectors. This information is then used to modulate a base VSR model, enabling adaptive handling of various compression levels. The model further utilizes metadata like residual maps for accurate frame alignment, enhancing the bidirectional recurrent network's performance. In addition to the aforementioned multi-frame-based models, VCISR \cite{wang2023vcisr} introduces an approach for the blind single image super-resolution (SISR) task that focuses on enhancing single-frame input affected by video compression artifacts, relying solely on spatial information.

\section{Preliminary}
Before delving into the specifics of our work, it is important to discuss the inspirations behind this idea. Deep generative models, such as Denoising Diffusion Probabilistic Models (DDPMs) \cite{Ho2020}, offer a compelling alternative to GANs \cite{goodfellow2014generativeadversarialnetworks,dehaqi2021adversarial} without the need for adversarial training, careful optimization, or the risk of missing parts of the data distribution. DDPMs achieve this by training denoising models to progressively transform Gaussian noise into images through a Markov chain process, providing a stable and effective generative approach. While DDPMs produce high-quality images through a lengthy generative process, this method requires numerous iterations, making it significantly slower than GANs. For instance, generating 50,000 images of size 256 × 256 can take nearly 1,000 hours on a Nvidia 2080 Ti GPU, which becomes increasingly impractical for larger resolutions \cite{song2021denoising}. To address this efficiency gap, Denoising Diffusion Implicit Models (DDIMs) \cite{song2021denoising} were introduced as a more efficient alternative to DDPMs. DDIMs generalize the forward Diffusion process from the Markovian framework used in DDPMs to non-Markovian processes. This allows for the creation of "short" generative Markov chains that can produce high-quality samples in far fewer steps. Cold Diffusion \cite{bansal2023cold} further explores the boundaries of Diffusion models by eliminating the reliance on Gaussian noise or randomness altogether. Instead of using noise, it leverages arbitrary image transformations and degradations, training a restoration network to reverse these transformations. This approach challenges the traditional theoretical frameworks of Diffusion models and opens the door to new types of generative models with distinct properties compared to conventional methods. Moreover, the use of Gaussian noise schedules not only prevents Stable Diffusion models \cite{Rombach2022} from generating images with mean brightness greater or less than 0 (on a scale of -1 to 1), but also proves to be an overextension of model's capacity. This is particularly true for restoration tasks, where the model must remove both artificially added Gaussian noise and existing artifacts \cite{10377127}. Since compressed frames are not a natural intermediate step in the vanilla Diffusion process, the restoration process does not need to start from pure noise, nor does it require a large number of inference steps or a large model size—advantages that are critical for real-world applications. With all this in mind, we present \emph{DiQP} as a proof of concept for using Denoising Diffusion to directly address the complex artifacts introduced by video compression in 8K resolution without adding artificial Gaussian noise \cite{Zhou2022}. 

\section{Methodology}
{\bf Big Picture:} \emph{DiQP} consists of 4 different parts which are explained in depth in this section. But first let’s revisit the problem. Let \begin{math}F_{\text{raw}} \in \mathbb{R}^{T \times H \times W \times C}\end{math}, be a sequence of raw target frames without added artifacts and distortions. \begin{math} T, H, W, C \end{math} are the frame number, height, width and channel number, respectively. Now  if we consider  \begin{math} CODEC_\text{Decode}(CODEC_\text{Encode}(F_{\text{raw}},QP)) = F_{\text{qp}} = F_{\text{raw}} + \text{noise}_{\text{qp}} \end{math}, our model aims to predict the \begin{math} \text{noise}_{\text{qp}} \end{math} as accurately as possible. Therefore, we formulate our model as \begin{math} DiQP(F_{\text{iqp}},QP,Z) = \text{noise}_{\text{qp}}' \end{math}, where \begin{math} F_{\text{iqp}} \end{math} is a randomly selected window from the original 8K frame, calculated by the Hadamard product of  \begin{math} F_{\text{qp}} \end{math} and a random binary mask \begin{math} M^{h \times w }\end{math} : \begin{math} F_{\text{iqp}} = F_{\text{qp}} \circ M \end{math}. The reconstructed output frame sequence \begin{math} F_{\text{res}}\end{math} is then obtained as \begin{math} F_{\text{res}} = F_{\text{iqp}} + \text{noise}_{\text{qp}}' \end{math}.  Additional input Z includes both conditional inputs and the inputs specifically for the Look Ahead and Look Around modules, which are discussed in detail later in this section.

For a fair comparison with existing methods, we use the commonly adopted Charbonnier loss \cite{Charbonnier1994} between the reconstructed frame sequence and the ground truth or raw sequence, defined as: 
\begin{equation} \mathcal{L}=\sqrt{\|F_{\text{res}} - (F_{\text{raw}} \circ M)\|^2 + \epsilon^2} \end{equation} 
and \begin{math}\epsilon = 10^{-3}\end{math} is a constant in all the experiments.
\begin{figure}[!htb]
  \centering
  \includegraphics[width=0.9\linewidth]{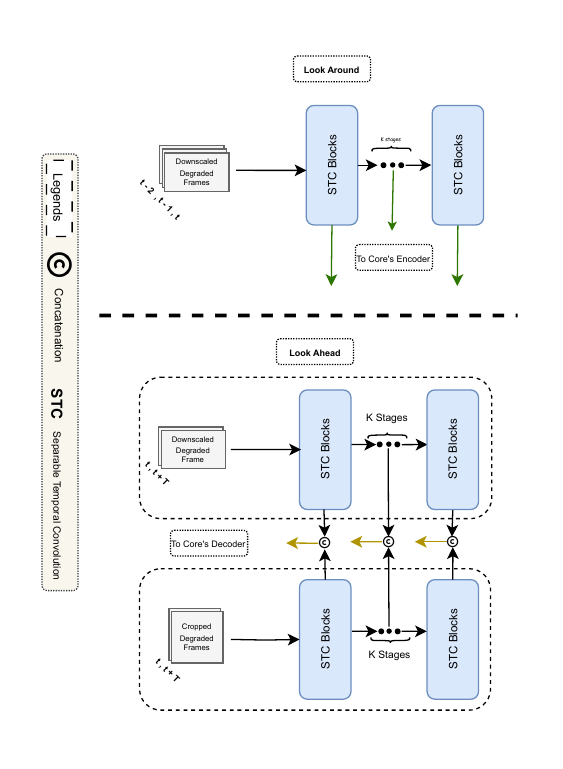}
  \caption{The Architectures of Look Around and Look Ahead Models}
  \label{fig:lar}
\end{figure}

\begin{figure*}[!htb]
    \centering
    \begin{subfigure}{0.32\textwidth}
        \centering
        \includegraphics[width=\linewidth]{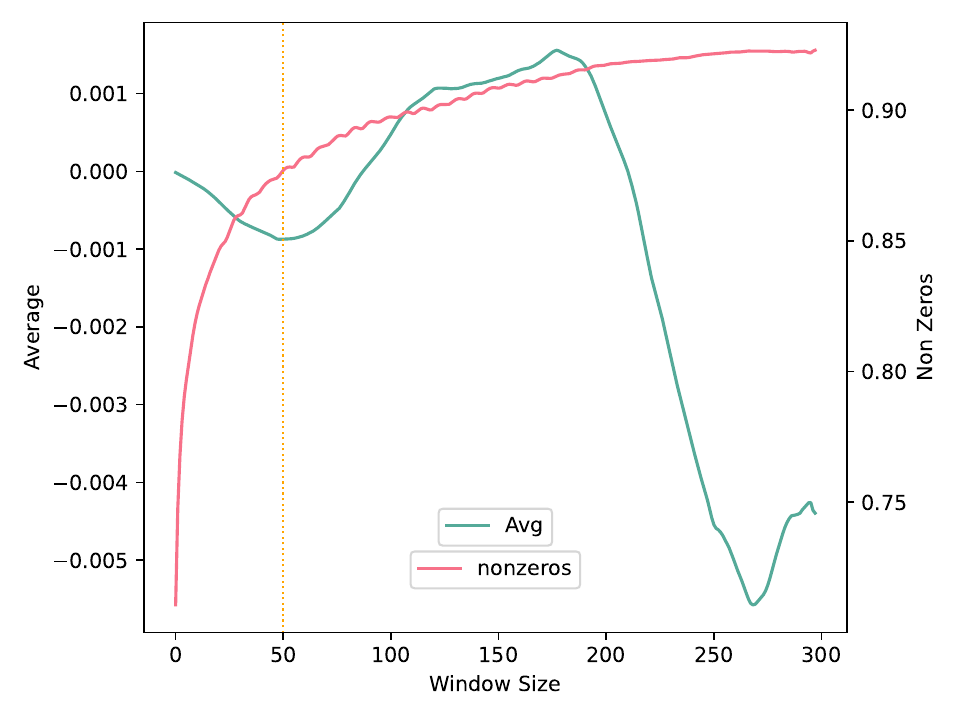}
        \caption{Average Difference vs. Non-Zero Pixels}
        \label{fig:avgnon}
    \end{subfigure}
    \hfill
    \begin{subfigure}{0.32\textwidth}
        \centering
        \includegraphics[width=\linewidth]{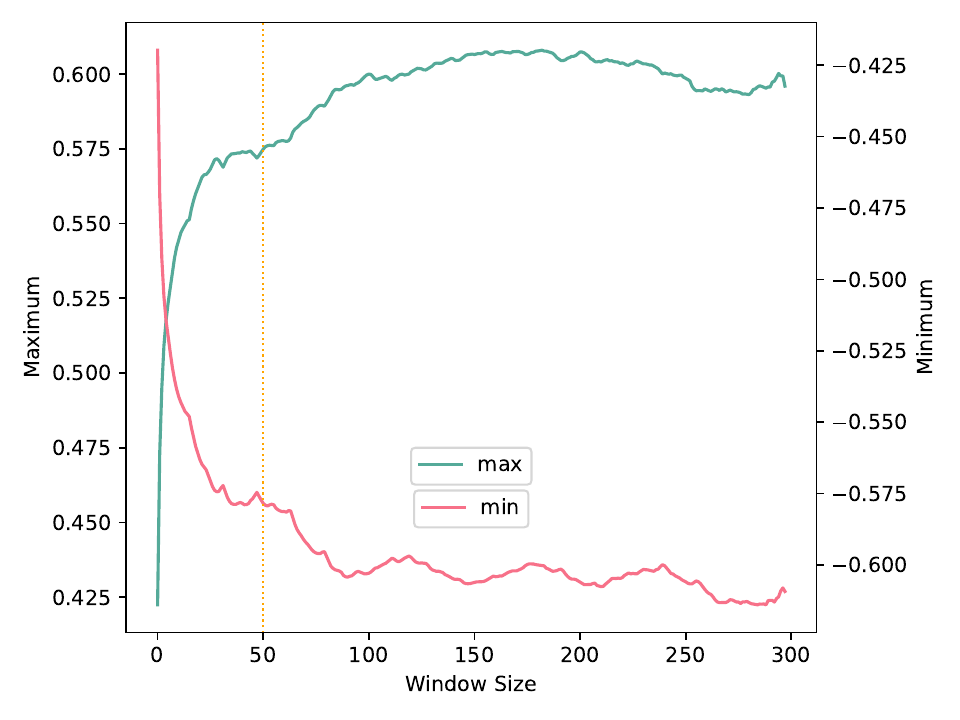}
        \caption{Maximum vs. Minimum Difference}
        \label{fig:minmax}
    \end{subfigure}
    \hfill
    \begin{subfigure}{0.32\textwidth}
        \centering
        \includegraphics[width=\linewidth]{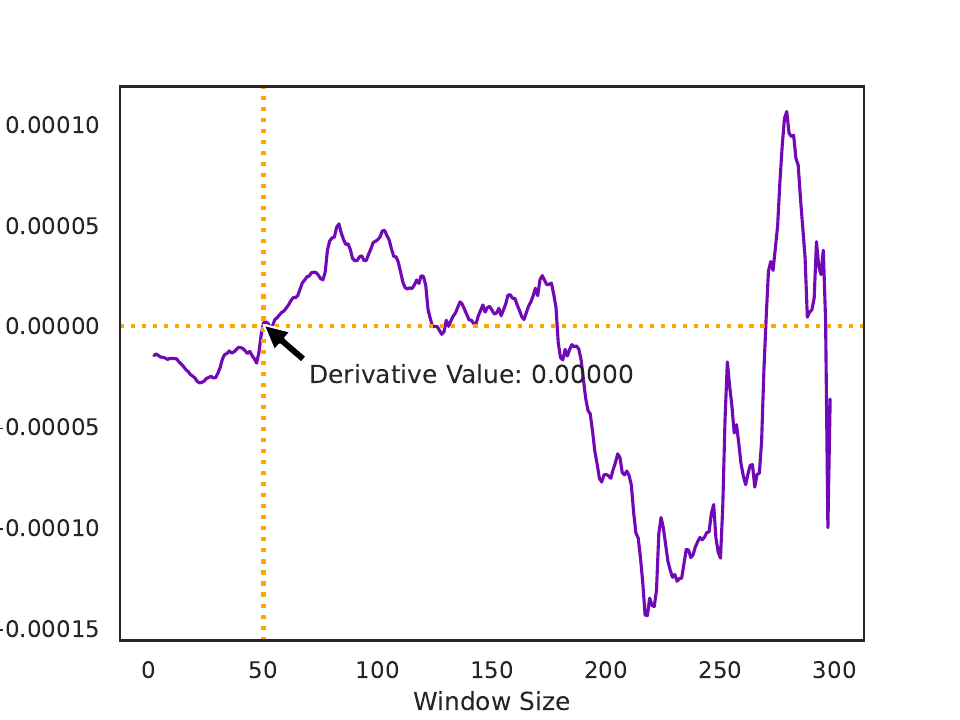}
        \caption{First Derivative of the Average Difference per Window Size}
        \label{fig:derivative}
    \end{subfigure}
    
    \caption{Frame difference analysis reveals the most significant changes at a temporal window size (T) of 50, indicating it as the optimal size for the Look Ahead model.}
    \label{fig:future}
\end{figure*}

\subsection{Look Around}
Due to the large size of 8K frames (7680 * 4320) and the limited coverage of the model's randomly selected (512 * 512) window which is less than 1\% of the whole frame, we introduce a supplementary model called \emph{Look Around} (Fig. \ref{fig:lar}) to provide the Core model with a better understanding of the surrounding context. To achieve this, we feed the \emph{Look Around} model, which consists of Temporal Separable CNN \cite{xie2018rethinking} blocks, with bicubically down sampled ($DS$) versions of the original 8K frames. The extracted features from this model are then used to guide the Core model. The Look Around model consists of K blocks of these temporal CNNs, with the output of each block being added separately to the corresponding level of the Core's Encoder part. This addition helps the Encoder better understand the input. The entire process can be formulated as follows:

\begin{equation}
LookAround(DS(F_{\text{qp}}^{7680 \times 4320})) = L_{\text{ar}}
 \end{equation} 

\subsection{Look Ahead}

Since our model employs a sliding window-based method which is not easily scalable to long sequence modeling \cite{Liang2024} and due to the challenges associated with estimating optical flow, particularly in highly compressed videos where inaccurate predictions and increased computational overhead can negatively impact the performance \cite{YingweiWang2023}, we introduce an auxiliary model called \emph{Look Ahead} (Fig. \ref{fig:lar}) to address this limitation and enhance the model's ability to anticipate future events and changes. The Look Ahead model takes the down-sampled of next T frame from the last frame in the input sequence and extracts informative features for the Core model. In addition to that frame, the Look Ahead model is also fed the same window coordination of input from the future frame. These two groups of data are processed separately and then concatenated ($\oplus$). Unlike the Look Around model, these extracted features are added to the corresponding levels of the Core's Decoder part. This addition enhances the decoder's restoration abilities. Furthermore, a weight decay factor ($WDF$) is incorporated to control the influence of the Look Ahead model. This decay factor proves particularly beneficial when processing the last T frames of the clip, as the last frame is used as input for these frames. The entire process can be formulated as follows: 
Let's denote the input frame set as \begin{math}\{F_{\text{qp1}}, F_{\text{qp2}}, \ldots, F_{\text{qpn}}\}\end{math} where \begin{math}F_{\text{qpn}}\end{math} represents the last frame in the input set. The frame of interest is then \begin{math}F_{\text{qp(n+T)}}\end{math} 
\begin{equation}
\begin{split}
    WDF &= \frac{\text{Total Frame Numbers} - \text{Middle Frame Position}}{\text{Total Frame Numbers}}
\end{split}
\end{equation}

\begin{equation}
\begin{split}
    &\text{LookAhead} \Big( 
        \underbrace{DS(F_{\text{qpn}}^{7680 \times 4320}) \oplus DS(F_{\text{qp(n+T)}}^{7680 \times 4320})}_{\text{Group 1}}, \\
    &\qquad \underbrace{(F_{\text{iqpn}} \oplus F_{\text{qp(n+t)}} \circ M)}_{\text{Group 2}}
    \Big) \times WDF = L_{\text{ah}}
\end{split}
\end{equation}

To determine the optimal temporal window size (T) for the Look Ahead model, we conducted an experiment analyzing how the input changes as the window size varies. Specifically, we randomly selected a frame, referred to as N, and subtracted it from each subsequent frame, up to the last frame (N+1 to frame 300). For each subtraction result, we calculated the minimum, maximum, total number of non-zero pixels, and the average. This process was repeated for various window sizes to evaluate the differences and identify the optimal temporal resolution.

Our results, as shown in Fig. \ref{fig:future}, indicate that the most significant changes occur at a window size of 50. The magnitude of change between window sizes 1 to 50 is considerably greater than that of between 50 to 299. Furthermore, the first derivative of the mean change, also depicted in Fig. \ref{fig:derivative}, approaches zero around window size 50.

\begin{figure}[H]
  \centering
  \includegraphics[width=0.9\linewidth]{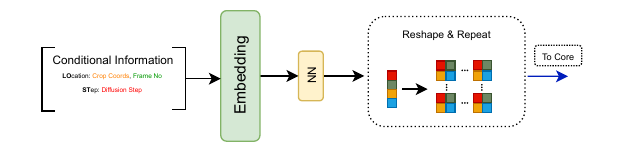}
  \caption{LOST embedding}
  \label{fig:lost}
\end{figure}

\subsection{LOST embedding}
When utilizing Denoising Diffusion for restoration tasks, a key challenge lies in effectively incorporating both degraded data and additional conditional data such as Diffusion time steps into the model. An improved conditional framework can significantly enhance the generative potential of Denoising Diffusions, guiding them towards producing realistic output that accurately matches the original sources \cite{YiZhang2023}. Thus, we employ an alternative approach to fully leverage the capabilities of the Diffusion model.

We introduce two types of conditional data to our model: 1) {\bf LO}cation and 2) Diffusion {\bf ST}ep, collectively referred to as \emph{LOST}! (Fig. \ref{fig:lost}). Location data encompasses the index of the middle frame within the entire clip, as well as the height and width of the input window in both original and down scaled sizes (used in Look Ahead and Look Around). For instance, the scaled crop points can be calculated as:
\begin{equation}
\begin{split}
        \text{Scale}_{d} &= \frac{\text{WindowSize}_{d}}{\text{OriginalSize}_{d}}, \\
    \text{CropPoint}_{\text{Downscaled}, d} &= \left\lfloor \text{CropPoint}_{\text{Original}, d} \times \text{Scale}_{d} \right\rfloor, \\
    & \quad \quad \quad \quad d \in \{\text{Width, Height}\}
\end{split}
\end{equation}

The step embedding is simply the Quantization Parameter (QP) used for video encoding. For each of these six values, we have a dedicated embedding that is trained independently. After obtaining and concatenating these embeddings into a larger vector, we pass the vector through a neural network (NN) with SiLU activation \cite{Elfwing2018} to produce a more informative and compact embedding. This embedding is then concatenated with the output of each block, serving as guidance and conditioning for subsequent blocks to optimize the model's performance. However, since the NN output is a vector, we reshape it into a matrix with dimensions matching the core's kernel size and replicate it horizontally and vertically to align with the corresponding block sizes (Algorithm \ref{alg:lost}).
\begin{algorithm}
\caption{LOST}\label{alg:lost}
\begin{algorithmic}[1]
\small
\REQUIRE Conditional information (step, frame number, window starting point scaled and original) and Size of the final output
\ENSURE Processed and encoded location and step information

\STATE Apply embedding function to all conditional information
\STATE Concatenate all embedded tensors on the last dimension
\STATE Apply the NN on the embedded tensors
\STATE Reshape the output from shape $(1, L)$ to $(1, l, l)$ where $l = \sqrt{\text{Size}}$
\STATE Repeat the reshaped tensor along new dimensions
\STATE Create a new dimension with size $k = \text{Size} / l$ by replicating each element in the first reshaped dimension and the second reshaped dimension $l$ times
\STATE The resulting tensor will have the shape $(1, k \times l, k \times l)$
\end{algorithmic}
\end{algorithm}

\subsection{Core Model}
The Core model is a U-shaped hierarchical network with skip connections between the encoder and the decoder. To be specific, given a triplet of degraded frames \begin{math}F_{\text{iqp}} \in \mathbb{R}^{3 \times H \times W \times C}\end{math}, the Core model first applies a 3D convolutional layer with LeakyReLU and a kernel size of 3 to extract low-level features. Next, following the design of the U-shaped structures \cite{Isola2017,Ronneberger2015}, the feature maps are passed through K encoder stages. Each stage contains a stack of the Transformer blocks and one down-sampling layer. The output of each stage is then concatenated with the output of $K$-th layer of Look Around and LOST before going through down sampling. In the down-sampling layer, we first reshape the flattened features into 3D spatial-temporal feature maps, and then down-sample the maps. Then, a bottleneck stage with a stack of Transformer blocks is added at the end of the encoder. In this stage, only LOST is concatenated with the output of each block. For feature reconstruction, the decoder also contains K stages. Each consists of an up-sampling layer and a stack of Transformer blocks similar to the encoder. After that, the features input to the Transformer blocks are the concatenation of the up-sampled features and the corresponding features from the encoder through skip-connection and the output of $K$-th layer of the Look Ahead model and LOST. Next, the Transformer blocks are utilized to learn to restore the frames. After the K decoder stages, we reshape the flattened features to 3D feature maps and apply a 3D convolution layer with kernel size of 3 to obtain added artifacts and distortions to remove from the frames. Due to the high computational cost of the standard Transformer architecture \cite{Vaswani2017,Dosovitskiy2021} and its limitations in capturing local dependencies \cite{Wu2021,li2021localvit}, we created a spatio-temporal compatible Transformer block based on the Locally enhanced Window (LeWin) Transformer introduced by \cite{Wang2022}. This block benefits from two key designs: Window-based Multiheaded Self-Attention, which performs self-attention within non-overlapping local windows, significantly reducing computational cost, and an enhanced Feed-Forward Network that leverages local context.

\begin{figure*}[ht]
  \centering
  \begin{subfigure}[ht]{\textwidth}
    \centering
    \includegraphics[width=0.8\textwidth]{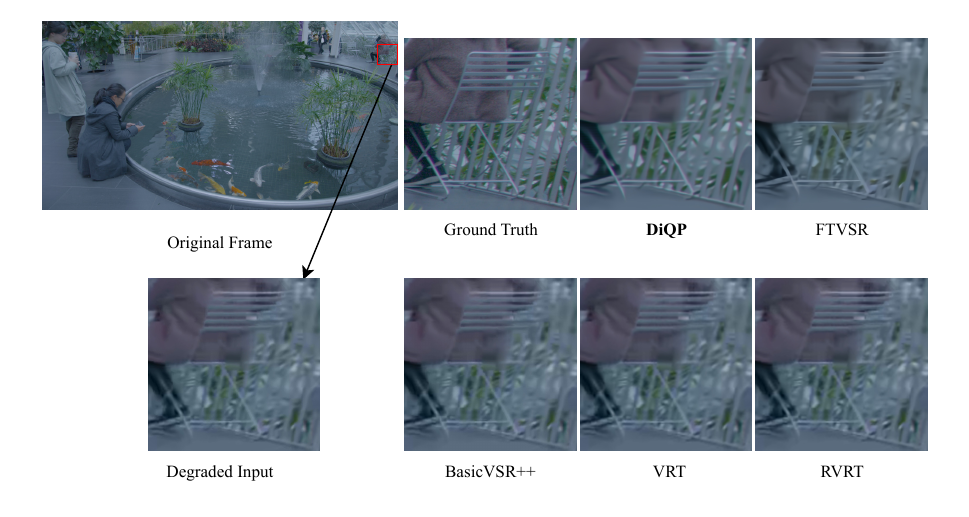}
    \caption{Comparison of visual quality produced by different methods on sequence 5 of the SEPE8K dataset.}
    \label{fig:resfive}
  \end{subfigure}
  \vspace{1em} 
  \begin{subfigure}[ht]{\textwidth}
    \centering
    \includegraphics[width=0.8\textwidth]{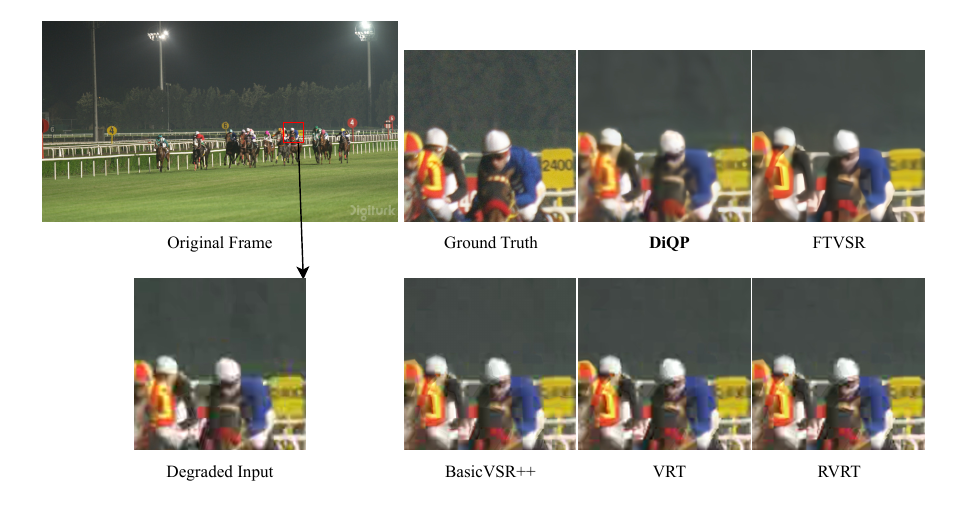}
    \caption{Comparison of visual quality produced by different methods on the Race Night sequence of the UVG dataset.}
    \label{fig:resrace}
  \end{subfigure}
  \caption{Visual comparisons of different methods on SEPE8K and UVG datasets. DiQP outperforms the baseline models in both datasets.}
  \label{fig:combined}
\end{figure*}

\section{Experiments}
\paragraph{Datasets:} We used the SEPE8K dataset \cite{10.1145/3587819.3592560} for our training. This dataset comprises 40 different 8K (8192 x 4320) video sequences, each captured at a framerate of 29.97 frames per second (FPS) with a duration of 10 seconds. We randomly split the dataset into 30, 5, and 5 sequences for training, testing \footnote{Sequences 5, 11, 21, 23, 26}, and validation, respectively. Using \emph{ffmpeg} with the help of \emph{NVIDIA A6000 Ada} GPU, we created frames from encoded videos using two codecs, HEVC/H.265 and AV1, with varying Quantization Parameters (QPs). For HEVC, we used QPs ranging from 3 to 51 (maximum) with a step size of 3, resulting in 17 quality levels. For AV1, we used QPs from 3 to 255 (maximum) with the same step size, yielding 85 quality levels. The total data occupied approximately 40 TB of storage. For training, we divided each video into 100 non-overlapping segments, each containing three frames. After loading the frames, we randomly selected 512*512 non-tile-wise window crops to prevent probable boundary artifacts. To broaden the evaluation of our model and ensure a fair comparison, we also tested it on the UVG 4K dataset \cite{Mercat2020}, specifically selecting videos with a duration of 12 seconds, given the very limited availability of 8K datasets. 

\paragraph{Implementation Details:} 
Due to the performance gap between HEVC and AV1 in the high-resolution domain \cite{Yuan2023}, we trained the same model on each codec separately. Training was conducted on a server with 8 \emph{NVIDIA A100} GPUs, taking 40 epochs for AV1 and 200 epochs for HEVC. The total training time, including experiments for the ablation study, was 40 days.

Following the common training strategy for Transformers \cite{Vaswani2017}, we employed the AdamW optimizer \cite{Loshchilov2017} with momentum terms of (0.9, 0.999) and a weight decay of 0.02. We also applied a learning rate warmup \cite{Goyal2017} for approximately 3\% of the initial epochs.

\paragraph{Evaluation Metrics:}
We adopt the commonly-used PSNR and SSIM \cite{article} metrics to evaluate the restoration performance. These metrics are calculated in the RGB.

\begin{table}[ht] 

\centering
\caption{Performance Comparison on SEPE8K and UVG with AV1 (QP = 255) and HEVC (QP = 51) codecs}
\begin{tabular}{c}
    \begin{subtable}[t]{0.9\linewidth}
        \centering
        \caption{SEPE8K with AV1 (QP = 255)}
        \begin{tabular}{lcc}
        \toprule
        \textbf{Model} & \textbf{PSNR} & \textbf{SSIM} \\
        \midrule
        DiQP            & 34.868 & 0.8611 \\
        FTVSR           & 33.201 & 0.8525 \\
        RVRT            & 33.113 & 0.8577 \\
        VRT             & 33.101 & 0.8501 \\
        BasicVSR++      & 33.105 & 0.8469 \\
        Degraded Input  & 33.095 & 0.8463 \\
        \bottomrule
        \end{tabular}
        \label{tab:sepe8k_av1}
    \end{subtable} \\[1em]
    \begin{subtable}[t]{0.9\linewidth}
        \centering
        \caption{SEPE8K with HEVC (QP = 51)}
        \begin{tabular}{lcc}
        \toprule
        \textbf{Model} & \textbf{PSNR} & \textbf{SSIM} \\
        \midrule
        DiQP            & 34.197 & 0.8538 \\
        FTVSR           & 32.316 & 0.8472 \\
        RVRT            & 32.281 & 0.8491 \\
        VRT             & 32.213 & 0.8464 \\
        BasicVSR++      & 32.211 & 0.8395 \\
        Degraded Input  & 32.206 & 0.8393 \\
        \bottomrule
        \end{tabular}
        \label{tab:sepe8k_hevc}
    \end{subtable} \\[1em]
    \begin{subtable}[t]{0.9\linewidth}
        \centering
        \caption{UVG with AV1 (QP = 255)}
        \begin{tabular}{lcc}
        \toprule
        \textbf{Model} & \textbf{PSNR} & \textbf{SSIM} \\
        \midrule
        DiQP            & 32.551 & 0.8662 \\
        FTVSR           & 31.880 & 0.8511 \\
        RVRT            & 31.711 & 0.8545 \\
        VRT             & 31.708 & 0.8506 \\
        BasicVSR++      & 31.706 & 0.8463 \\
        Degraded Input  & 31.702 & 0.8461 \\
        \bottomrule
        \end{tabular}
        \label{tab:uvg_av1}
    \end{subtable} \\[1em]
    \begin{subtable}[t]{0.9\linewidth}
        \centering
        \caption{UVG with HEVC (QP = 51)}
        \begin{tabular}{lcc}
        \toprule
        \textbf{Model} & \textbf{PSNR} & \textbf{SSIM} \\
        \midrule
        DiQP            & 31.965 & 0.8590 \\
        FTVSR           & 31.289 & 0.8485 \\
        RVRT            & 31.271 & 0.8498 \\
        VRT             & 31.270 & 0.8481 \\
        BasicVSR++      & 31.269 & 0.8430 \\
        Degraded Input  & 31.267 & 0.8426 \\
        \bottomrule
        \end{tabular}
        \label{tab:uvg_hevc}
    \end{subtable}
\end{tabular}
\label{tab:res}
\end{table}

\subsection{Comparison}
We selected the four representative methods in video restoration (VRT \cite{Liang2024}, RVRT \cite{Liang2022}, BasicVSR++ \cite{Chan2022},  and FTVSR) as baselines to compare our model. We present quantitive comparison results between \emph{DiQP} and baselines in Table \ref{tab:res}. The test was conducted with the maximum QP available for both codecs. To develop a better understanding, we also added the metrics for degraded input.
As we can see, \emph{DiQP} achieves the best performance on SEPE8K and UVG for both codecs. Compared with the baseline models, it improves the PSNR  by significant margins of {\bf1.77 to 1.99dB} in SEPE8K and {\bf0.84 to 0.69dB} in UVG. For comparison with UVG, due to the fixed dimensions of the LOST embedding (learned specifically for 8K domains), we had to upsample the UVG 4K frames to 8K using bicubic interpolation before performing restoration. After restoration, we downscaled the results and compared them with the original raw frames. This process likely affected the overall results, as some fine details may have been lost during the upscaling and downscaling steps. In Table \ref{tab:params_runtime_comparison} we present a comparison of the model parameters and runtime across different methods, highlighting that \emph{DiQP}, despite having the highest number of parameters, achieves the fastest runtime. The visual comparisons of different methods shown in Fig. \ref{fig:resfive} and Fig. \ref{fig:resrace} indicate that \emph{DiQP} generates smoother and more clear HQ frames with removed artifacts, while other methods fail to restore fine textures and details. The second best performing model here is FTVSR because it has a better understanding of compression side effects on video. Due to limited space, additional results are provided in the supplementary material.

\begin{figure}[ht]
  \centering
  \includegraphics[width=0.9\linewidth]{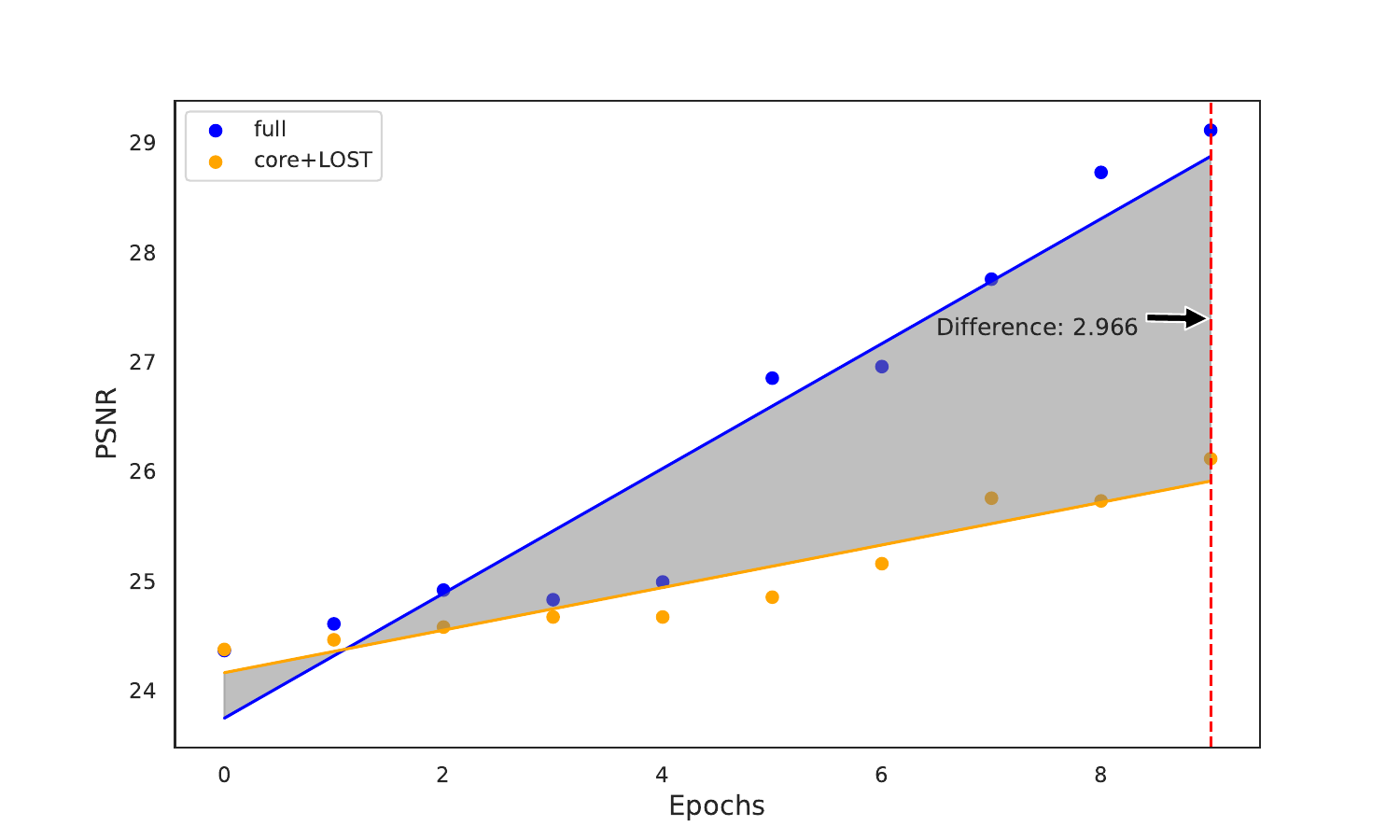}
  \caption{PSNR comparison after 10 epochs between the full DiQP model and a version without Look Ahead and Look Around modules, showing a 3dB gain with the full model.}
  \label{fig:ablation}
\end{figure}

\subsection{Ablation Study}
\label{sec:ablation}

\paragraph{Understanding the Role of Auxiliary Models} 
We conducted an ablation study to evaluate the impact of Look Ahead and Look Around models on the overall performance. Due to computational constraints, we focused our analysis on comparing our complete, fully-featured \emph{DiQP} model with a simplified version lacking the Look Ahead and Look Around modules. This targeted comparison allowed us to isolate the contributions of these two modules and better understand their role in achieving the final performance of the complete model.
In this experiment, both models were trained under identical conditions for 10 epochs. We then analyzed their output quality by calculating the PSNR between the generated results and the ground truth. Notably, after 10 epochs, we observed a significant difference of approximately {\bf 3dB}  in PSNR between the two models as illustrated in Fig. \ref{fig:ablation}.

\begin{table}[ht]
    \centering
    \caption{Model Parameters and Runtime Comparison}
    \begin{tabular}{lcc}
    \toprule
    \textbf{Method} & \textbf{Params (M)} & \textbf{Runtime (ms)} \\
    \midrule
    DiQP            & 79.37 & {\bf 125} \\
    FTVSR           & 10.8  & 499 \\
    RVRT            & 10.8  & 321 \\
    VRT             & 35.6  & 427 \\
    BasicVSR++      & {\bf 7.3}   & 135 \\
    \bottomrule
    \end{tabular}
    \label{tab:params_runtime_comparison}
\end{table}

\section{Conclusion} 
\emph{DiQP} is a novel Transformer-Diffusion model for 8K video restoration; specifically addressing the complex artifacts introduced by codec compression. By viewing the restoration process itself as a Deonising Diffusion model and leveraging the Quantization Parameter (QP) as the Diffusion step, we successfully applied this powerful framework to the challenging task of video restoration. Our approach demonstrates superior performance in restoring high-resolution videos from heavily compressed sources. The experimental results highlight the effectiveness of our model in recovering fine details and improving overall visual quality compared to other existing models. For future work, we plan to work on a more compatible LOST embedding since it is currently designed only for 8K resolution, and adapting it for other resolutions is challenging. 
{\small
\bibliographystyle{ieee_fullname}
\bibliography{export.bib}
}
\newcommand{\beginsupplement}{
    \setcounter{table}{0}
    \renewcommand{\thetable}{S\arabic{table}}
    \setcounter{figure}{0}
    \renewcommand{\thefigure}{S\arabic{figure}}
    \setcounter{section}{0}
    \renewcommand{\thesection}{S\arabic{section}}
}

\newpage
\section*{Supplementary Material}
\begin{figure*}[!htb]
  \centering
  \includegraphics[width=0.7\linewidth,height=0.9\textheight]{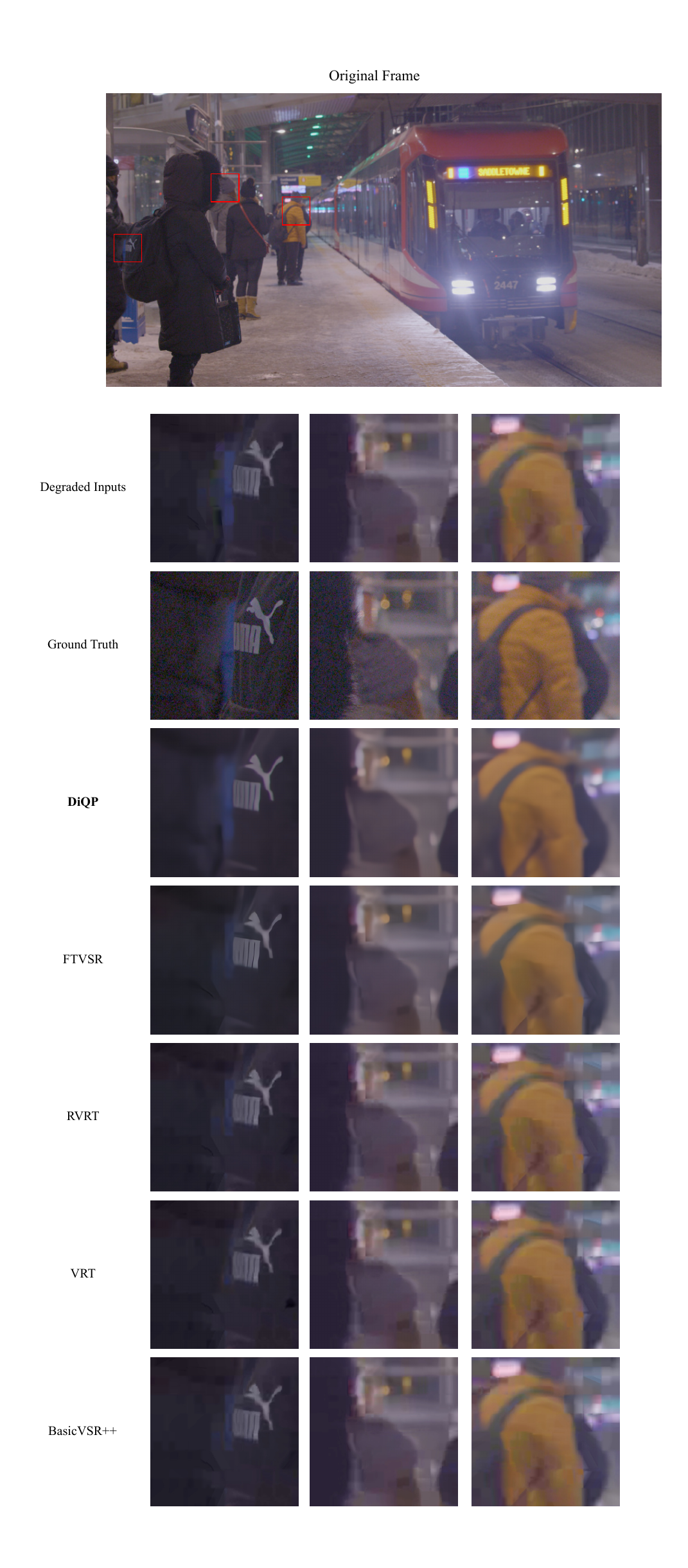}
  \caption{Visual comparisons of different methods on sequence 26 of SEPE8K encoded with QP=255 on AV1}
  \label{fig:26}
\end{figure*}

\beginsupplement
In this section, we delve deeper into the quantitative and qualitative results obtained from our model. Fig. \ref{fig:26} showcases frame 179 from sequence 26 of the SEPE8K dataset, along with our model's output. To better illustrate our model's performance, we highlight three specific crops from this frame: 1) the in-focus "bag" crop, demonstrating the model's ability to handle sharp details, 2) the out-of-focus "hat" crop with its intricate winter hat texture, testing the model's performance on challenging patterns, and 3) the out-of-focus "jacket" crop, further evaluating the model's ability to reconstruct detailed textures in less-than-ideal conditions. Our findings highlight that our model consistently delivers high-quality results in both in-focus and out-of-focus scenarios, effectively handling various textures and sharpness levels, unlike other models that struggle in such diverse situations.

To validate our model's robustness, we evaluated its performance under a range of compression levels (QPs). This included not only the most challenging highest QP scenario but also lower QPs to ensure comprehensive learning. As detailed in Tables \ref{tab:av1_psnr}, \ref{tab:av1_ssim}, \ref{tab:hevc_psnr}, and \ref{tab:hevc_ssim}, our model consistently demonstrates strong performance across various compression levels, from the least to the most compressed videos. Intuitively, our \emph{DiQP} exhibits a notably stronger performance advantage at lower QPs, with a margin of approximately 5 dB in the SEPE8K dataset and around 2 dB in the UVG dataset. This shows a superior ability to leverage the additional information available in less compressed videos.

\begin{table}[!htb]
\centering
\caption{PSNR Performance of \emph{DiQP} on AV1 Codec with Various QPs}

\begin{tabular}{lcccc}
\toprule
\textbf{QP} & \textbf{Dataset} & \textbf{Output} & \textbf{Input} \\
\midrule
51  & SEPE8K & 42.171 & 37.544 \\
    & UVG    & 37.001 & 35.124 \\
\midrule
102 & SEPE8K  & 40.478 & 36.774 \\
    & UVG    & 35.772 & 34.286 \\
\midrule
153 & SEPE8K  & 38.762 & 35.816 \\
    & UVG    & 34.962 & 33.644 \\
\midrule
204 & SEPE8K  & 36.818 & 34.564 \\
    & UVG     & 33.871 & 32.740 \\
\bottomrule
\end{tabular}
\label{tab:av1_psnr}
\end{table}

\begin{table}[!htb]
\centering
\caption{SSIM Performance of \emph{DiQP} on AV1 Codec with Various QPs}

\begin{tabular}{lcccc}
\toprule
\textbf{QP} & \textbf{Dataset} & \textbf{Output} & \textbf{Input} \\
\midrule
51  & SEPE8K & 0.9304 & 0.9105 \\
    & UVG     & 0.9148 & 0.8981 \\
\midrule
102 & SEPE8K  & 0.9047 & 0.8886 \\
    & UVG     & 0.8924 & 0.8788 \\
\midrule
153 & SEPE8K  & 0.8852 & 0.8708 \\
    & UVG     & 0.8831 & 0.8699 \\
\midrule
204 & SEPE8K  & 0.8730 & 0.8584 \\
    & UVG     & 0.8731 & 0.8587 \\
\bottomrule
\end{tabular}
\label{tab:av1_ssim}
\end{table}

\begin{table}[!htb]
\centering
\caption{PSNR Performance of \emph{DiQP} on HEVC Codec with Various QPs}

\begin{tabular}{lcccc}
\toprule
\textbf{QP} & \textbf{Dataset} & \textbf{Output} & \textbf{Input} \\
\midrule
12  & SEPE8K  & 43.063 & 38.036 \\
    & UVG     & 38.396 & 36.018 \\
\midrule
33  & SEPE8K  & 38.099 & 35.808 \\
    & UVG     & 34.882 & 33.590 \\
\midrule
42  & SEPE8K  & 35.829 & 34.385 \\
    & UVG     & 33.660 & 32.665 \\
\bottomrule
\end{tabular}
\label{tab:hevc_psnr}
\end{table}

\begin{table}[!htb]
\centering
\caption{SSIM Performance of \emph{DiQP} on HEVC Codec with Various QPs}

\begin{tabular}{lcccc}
\toprule
\textbf{QP} & \textbf{Dataset} & \textbf{Output} & \textbf{Input} \\
\midrule
12  & SEPE8K  & 0.9456 & 0.9241 \\
    & UVG     & 0.9307 & 0.9133 \\
\midrule
33  & SEPE8K  & 0.8842 & 0.8704 \\
    & UVG     & 0.8835 & 0.8684 \\
\midrule
42  & SEPE8K  & 0.8695 & 0.8564 \\
    & UVG     & 0.8728 & 0.8571 \\
\bottomrule
\end{tabular}
\label{tab:hevc_ssim}
\end{table}

\end{document}